\documentclass[a4paper]{article}
\usepackage{url}
\usepackage{diagbox}
\usepackage{tabu, array}
\usepackage{multirow}
\usepackage{xcolor}

\usepackage{soul}

\usepackage{cite}
\usepackage{INTERSPEECH2022}

\title{VCSE: Time-Domain Visual-Contextual Speaker Extraction Network}
\name{Junjie Li$^1$, Meng Ge$^{1,2,*}$, Zexu Pan$^2$, Longbiao Wang$^{1,*}$, Jianwu Dang$^{1,3}$
\thanks{$^*$ Corresponding author.}
}

\address{
  $^1$ Tianjin Key Laboratory of Cognitive Computing and Application,\\ College of Intelligence and Computing, Tianjin University, Tianjin, China\\
  $^2$ Department of Electrical and Computer Engineering, National University of Singapore, Singapore\\
  $^3$ Japan Advanced Institute of Science and Technology, Ishikawa, Japan}

\email{\{mrjunjieli,gemeng,longbiao\_wang\}@tju.edu.cn, pan\_zexu@u.nus.edu}

\begin{document}
%
\maketitle
\begin{abstract}
Speaker extraction seeks to extract the target speech in a multi-talker scenario given an auxiliary reference. Such reference can be auditory, i.e., a pre-recorded speech,  visual, i.e., lip movements, or contextual, i.e., phonetic sequence. References in different modalities provide distinct and complementary information that could be fused to form top-down attention on the target speaker. Previous studies have introduced visual and contextual modalities in a single model.  In this paper, we propose a two-stage time-domain visual-contextual speaker extraction network named VCSE, which incorporates visual and self-enrolled contextual cues stage by stage to take full advantage of every modality. In the first stage, we pre-extract a target speech with visual cues and estimate the underlying phonetic sequence. In the second stage, we refine the pre-extracted target speech with the self-enrolled contextual cues. Experimental results on the real-world Lip Reading Sentences 3 (LRS3) database demonstrate that our proposed VCSE network consistently outperforms other state-of-the-art baselines.

\end{abstract}


\begin{figure*}[htbp] 
\centering 
\includegraphics[width=1\textwidth]{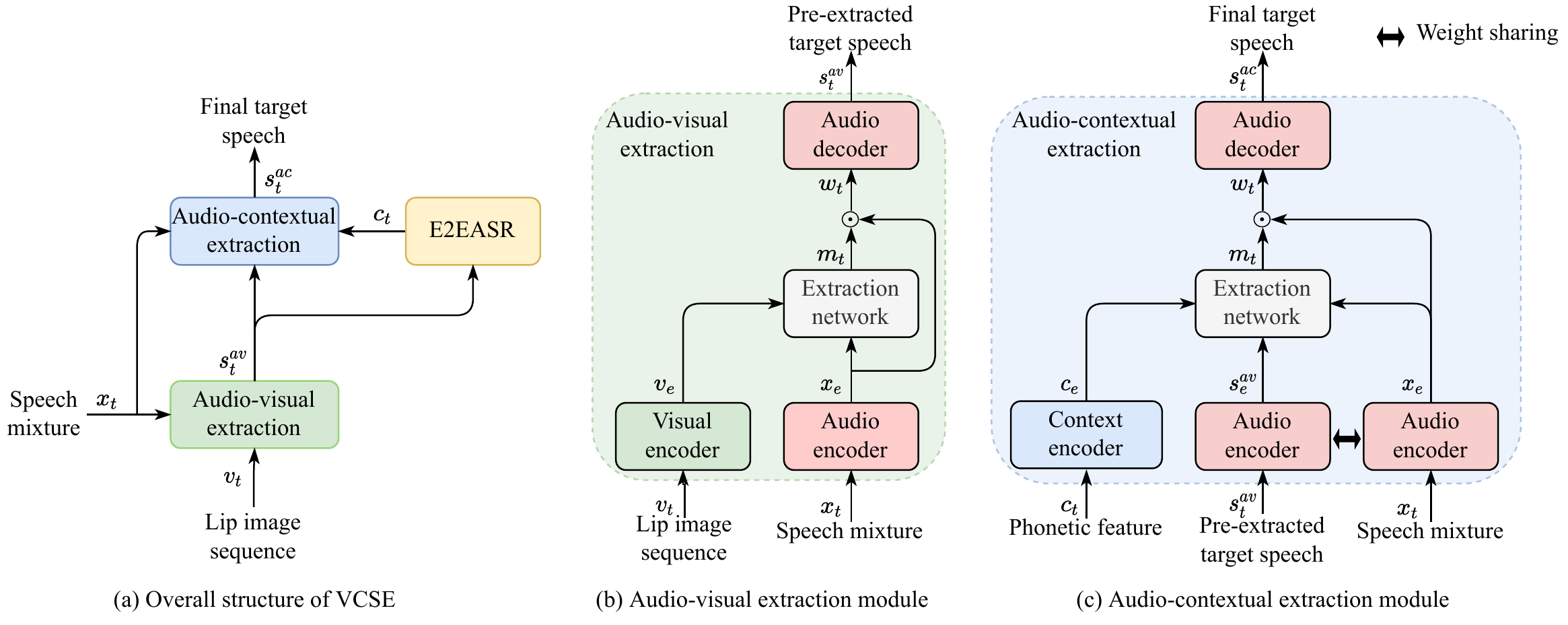}
\caption{(a)The overall structure of VCSE. $x_t, v_t, s_t^{av}, c_t$ and $s_t^{ac}$ represent time-domain speech mixture, lip image sequence, pre-extracted target speech, phonetic feature, and final target speech, respectively. (b) The structure of the audio-visual extraction module. $\odot$ denotes element-wise multiplication. (c) The structure of the audio-contextual extraction module. }
\label{fig:network}
\end{figure*}

\section{Introduction}
\label{sec:intro}

Speaker extraction aims to separate the speech of the target speaker from a multi-talker mixture signal, which is also known as the cocktail party problem \cite{bronkhorst2000cocktail}. This is a fundamental but crucial problem to solve in signal processing that benefits a wide range of downstream applications, such as hearing aids \cite{wang2008time}, active speaker detection~\cite{tao2021someone}, speaker localization~\cite{ge2022spex}, and automatic speech recognition (ASR) \cite{narayanan2014investigation,yoshioka2018multi}. Although human can easily do that, it is a huge challenge to realize it in machines.


Before the deep learning era, popular techniques were computational auditory scene analysis  \cite{wang2006computational} and non-negative matrix factorization  \cite{schmidt2006single}. The prior studies have laid the foundation for recent progress. With the advent and success of deep learning, speech separation algorithms such as permutation invariant training (PIT)  \cite{yu2017permutation}, deep clustering \cite{hershey2016deep}, wavesplit \cite{zeghidour2021wavesplit} and Conv-TasNet \cite{luo2019conv}, tackle the cocktail party problem by separating every speaker out in the mixture signal. Although a great success, there is inherent ambiguity in speaker labeling of the separated signals. An auxiliary reference, such as a pre-recorded speech signal \cite{ge2020spex+,han2020continuous,9067003,wang2018voicefilter} or video frame sequence \cite{reentry,aldeneh2021role,ephrat2018looking,seg_pan}, can be used to solve the speaker ambiguity. The speaker extraction algorithm employs such auxiliary reference to form top-down attention on the target speaker and extracts its speech.

Neuroscience studies \cite{remley2017neuroscience,ward2017enhanced} suggest that human perception is multimodal. According to the \textit{reentry} theory~\cite{edelman1987neural}, multimodal information is processed in our brain in an interactive manner, which educates each other.  At the cocktail party, we hear the voice of the person, observe the lip motions and understand the contextual relevance. The lip motions that are synchronized with the target speech help us better capture the places of articulation, and it is robust against acoustic noise. The contextual information connects the preceding or following parts of the speech, which helps fill up the current severely corrupted speech by inferring from the context. The information from different modalities complements each other and together forms effective communication~\cite{rosenblum2008speech,pan2020multi,chen2020multi,9755926}. Inspired by these prior studies, we aim to emulate such a human perception process, to utilize the visual and contextual cues.

A number of audio-visual speaker extraction works explore the visual and contextual information by using the viseme-phoneme mapping cues~\cite{wu2019time,pan2021muse,pan2021usev}. They encode the lip images into visemes using a visual encoder pre-trained on the lip reading task, in which each viseme maps to multiple phonemes. However, such phoneme information derived from visual images only is weak, and the network may be using more of the lip motions cues derived from the visemes.

There are also studies incorporating ASR derived phonemes to explore contextual information in the speech separation~\cite{takahashi2020improving,li2020listen,petridis2018end} algorithm. In \cite{takahashi2020improving}, the authors propose a two-stage method. The first stage is to obtain separated speech signals from a mixture signal with PIT. 
Then, they estimate the contextual embedding and guide the model to obtain the final speech in the second stage.
In \cite{li2020listen}, the authors use the speech mixture and visual cues to get contextual information. 
This work utilizes visual and contextual modalities for the first time. Both of these two works are frequency-domain methods. In this paper, we aim to find a solution in the time-domain, as time-domain methods usually outperform frequency-domain counterparts by avoiding the difficult phase estimation problem~\cite{luo2019conv}. Instead of incorporating multiple modalities in a single model, we aim to introduce one modality in each single stage.

Motivated by the previous works, we propose a two-stage time-domain visual-contextual speaker extraction (VCSE) network. The VCSE network is conditioned on auxiliary visual  reference only, but it makes use of both the visual lip movement cues and self-enrolled contextual cues in the extraction process. 
In the first stage, the network pre-extracts the target speech and estimates the underlying phonemes using a pre-trained ASR system. In the second stage, the pre-extracted speech is refined with the contextual cues encoded from the self-enrolled phonetic sequence. Experimental results on real-world Lip Reading Sentences 3 (LRS3) database \cite{afouras2018lrs3} show that our VCSE network consistently outperforms other state-of-the-art baselines.


\section{VCSE network}
\label{sec:model}



We design a two-stage network to utilize visual and contextual cues completely. In the first stage, visual cues is introduced to pre-extract the target speech and estimate the underlying phonemes, because it is robust against acoustic noise. In the second stage, with the self-enrolled contextual cues encoded from phonetic embedding, the network can refine the current speech frame by inferring from the preceding and following parts of pre-extracted speech frames.

\subsection{Network architecture}
The VCSE network consists of an audio-visual extraction module, an end-to-end ASR (E2EASR) module, and an audio-contextual extraction module, as depicted in Fig.~\ref{fig:network} (a). In the first stage, the pre-extracted speech $s_t^{av}$ is extracted with the help of visual cues in the audio-visual extraction module, which takes the time-domain speech mixture $x_t$ and lip image sequence $v_t$ as inputs. The E2EASR module encodes $s_t^{av}$ to produce phonetic embedding $c_t$. In the second stage, the audio-contextual module takes the phonetic embedding $c_t$, pre-extracted target speech $s_t^{av}$ and speech mixture $x_t$ as inputs to acquire final target speech $s_t^{ac}$.



\subsubsection{Audio-visual extraction module}
\label{sec:AV}
Audio-visual extraction module employs the visual cues, i.e., lip image sequences as reference to pre-extract the target speech from the speech mixture. 

The audio-visual extraction module shares a similar structure with AV-ConvTasNet proposed in \cite{wu2019time}, which consists of four parts: video encoder, audio encoder, extraction network and audio decoder, as shown in Fig.~\ref{fig:network}(b). The audio encoder and decoder perform 1D convolution and de-convolution, respectively. The extraction network is a stack of several temporal dilated convolutional blocks (TCN). The visual encoder consists of a 3D convolution layer and an 18-layer ResNet. Different from~\cite{wu2019time} where the visual encoder is pre-trained on the lip reading task to capture viseme-phoneme mapping cues, we do not pre-train the visual encoder. We join-train the visual encoder with the other parts to let the network decide on its own intrinsically what is the best visual representation.

Considering a time-domain audio mixture signal $x_t$ and lip image sequence $v_t$, the visual encoder encodes $v_t$ into visual embedding $v_e$, which is time-aligned with the audio signal at frame-level. Audio encoder transforms the speech mixture $x_t$ into latent embedding $x_e$. The extraction network takes $v_e$ and $x_e$ as inputs and estimates the mask $m_t$.
The $m_t$ is element-wise multiplied with the $x_e$ to obtain the latent representation of the pre-extracted target speech $w_t$.
The audio decoder renders pre-extracted target speech $s_t^{av}$ from $w_t$.

\subsubsection{E2EASR module}

\begin{figure}[t] 
\centering 
\includegraphics[width=0.4\textwidth]{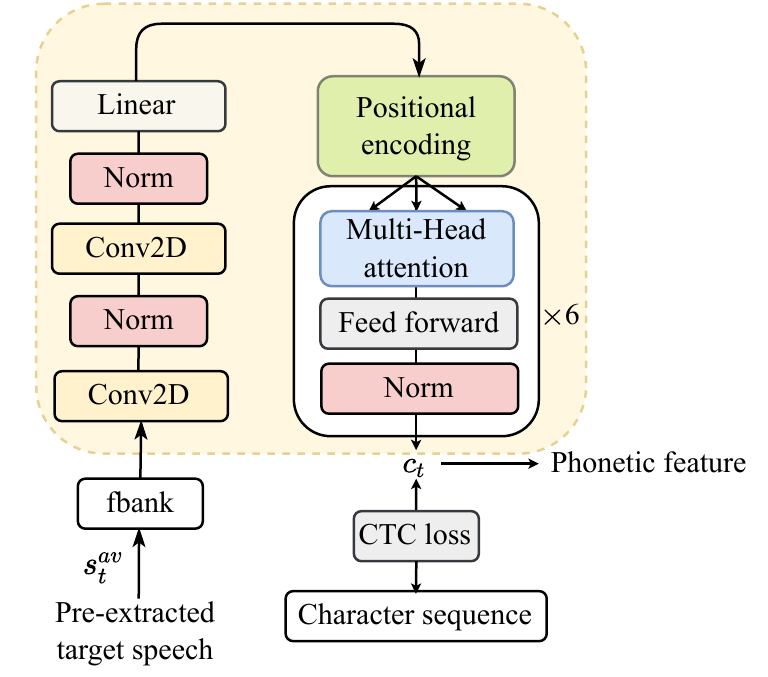}
\caption{Illustration of E2EASR module.} 
\label{fig:asr}
\vspace{-10pt}
\end{figure}

We adopt the encoder of  OpenTransformer network \footnote{https://github.com/ZhengkunTian/OpenTransformer} as our E2EASR module.  The OpenTransformer has a transformer architecture \cite{vaswani2017attention} and is trained with the connectionist temporal classification (CTC) loss function using a  clean speech signal, as depicted in Fig.~\ref{fig:asr}. 
The output of transformer encoder $c_t$ is the phonetic feature.

\subsubsection{Audio-contextual extraction module}
\label{sec:ac}


The audio-contextual extraction module has a similar architecture with the audio-visual  extraction module except for a context encoder replacing the visual encoder, as depicted in Fig.~\ref{fig:network}(c). Although sharing similar architecture, they do not share the model weights as the two modules perform different tasks, i.e., pre-extraction in the audio-visual extraction module and refinement in the audio-contextual extraction model. 

The context encoder is a stack of five 1D convolutional blocks with the exponential growth dilation factor $2^d$, where $d\in \{0,1,2,3,4\}$. A linear layer is followed to adjust the feature dimension. Each 1D conv block contains 1D conv with kernel size 5, channel size 256, and the exponential growth padding $\in \{2,4,8,16,32\}$, 1D batch normalization and ReLU activation function\footnote{https://github.com/pragyak412/Improving-Voice-Separation-by-Incorporating-End-To-End-Speech-Recognition}. 

The context encoder is designed to acquire the contextual embedding $c_e$ from the phonetic feature $c_t$. Two audio encoders encode two audio signals into the latent representations $s_e^{av}$, $x_e$,  and share the same weights. According to \cite{ge2020spex+}, it is helpful for separation to project the speech mixture and reference speech into the same latent space. These three features are concatenated and modeled by the extraction network to estimate mask $m_t$. The audio decoder decodes audio representation $w_t$ to final target speech $s_t^{ac}$.


\subsection{Loss function}
We adopt the scale-invariant signal-to-noise ratio (SI-SNR) \cite{le2019sdr,luo2019conv} as loss function to train the audio-visual extraction module and the audio-contextual extraction module in speech signal reconstruction, which is defined as:
\begin{equation}
    \left\{ \begin{array}{lr}
    s_{target}: = \frac{\hat{s}^\mathrm{T}s}{\Vert s \Vert ^2}s & \\
    e_{noise}: = \hat{s} -s_{target} & \\
    \mathcal{L}_{\text{SI-SNR}} (s, \hat{s}): = - 10log_{10} \frac{\Vert s_{target}\Vert^2 }{\Vert e_{noise} \Vert^2} &
    \end{array}
    \right.
\end{equation}
where $s$ and $\hat{s}$ denote the clean target speech and an estimated speech, respectively.

\subsection{Training strategy}

Our training process is divided into 5 steps. 

\begin{enumerate}
    \item We train the audio-visual extraction module alone with the SI-SNR loss function $\mathcal{L}_{\text{SI-SNR}} (s, s_t^{av})$.
    \item We pre-train the OpenTransformer network, which forms the E2EASR module in our VCSE network. The pre-trained OpenTransformer network reaches 9.5\% word error rate on the LRS3 test sets.
    \item We fix the weights of the audio-visual extraction module and the E2EASR module, and train the audio-contextual extraction module. It is worth noting that in this step of training, we use the phonetic feature $c_t$ encoded from the clean target speech instead of the pre-extracted speech. This enables the network to converge faster with such oracle phonetic feature. The SI-SNR loss function $\mathcal{L}_{\text{SI-SNR}} (s, s_t^{ac})$ is used in this step.
    \item We repeat the training in step 3, except that the phonetic feature $c_t$ in this step is encoded from the pre-extracted speech $s_t^{av}$. The SI-SNR loss function $\mathcal{L}_{\text{SI-SNR}} (s, s_t^{ac})$ is used in this step.
    \item We release all of the fixed module weights and fine-tune the whole system end-to-end. The SI-SNR loss function $\mathcal{L}_{\text{SI-SNR}} (s, s_t^{ac})$ is used in this step.
\end{enumerate}

\section{Experimental setup}
\label{sec:setup}
\subsection{Datasets}

In this paper, all modules are trained on the LRS3 \cite{afouras2018lrs3} dataset. This dataset contains  thousands of spoken sentences from TED and TEDx videos. There are 118,516 (252h), 31,982 (30h) and 1,321 (0.85h) utterances in pre-train, trainval and test sets, respectively, and there is no overlap between the videos used to create the test set and the ones used for the pre-train and trainval sets. The sampling rate of the audio signal is 16kHz.

\noindent 
\textbf{Automatic speech recognition:} The original  LRS3 pre-train set is used to pre-train the OpenTransformer network. We select utterances that are less than 20 seconds to pre-train the module to fit into GPU memory. 

\noindent
\textbf{Speaker extraction:} We simulate a two-speaker mixture sets to train and evaluate our VCSE network, using the LRS3 data-mix script \footnote{https://github.com/JusperLee/LRS3-For-Speech-Separation}. The long utterances are truncated to 3 seconds, and the utterances less than 3 seconds are dropped. The two-speaker mixtures are generated by selecting different utterances and mixed at various signal-to-noise ratios (SNR) between -5dB and 5dB. The corresponding video stream is sampled in 25 frames per second (fps). The lip region of each video frame is detected by face recognition algorithm, and we resize it to 120*120 pixels. 
The numbers of utterances for training, validation and test are 50,000, 5,000, 3,000 respectively.

\subsection{Training setup}

For audio-visual extraction module and audio-contextual extraction module,  adam is used as an optimizer.  And the initial learning rate is set to 0.001. Besides, the learning rate is halved if the validation loss increases consecutively for 3 epochs. The training process stops when validation loss increases consecutively for 6 epochs. 

For E2EASR module, we use CTC as loss function. We select adam as an optimizer. Warmup strategy \cite{vaswani2017attention} is adopted to adjust learning rate, which first increases the learning rate linearly and then decreases thereafter proportionally to the inverse square root of the step number.
The length of each phonetic embedding is 73, and the dimension of channel is 256.

\renewcommand{\arraystretch}{2}
\begin{table}[tp]
	\centering
	\fontsize{7}{8}\selectfont
	\caption{Results of various models using visual information (V), speaker embedding ($A_S$) or contextual information (C) as reference in terms of SI-SNRi (dB) and SDRi (dB). ``BSS'' and ``SE'' denote blind source separation and speaker extraction, respectively.  ``Reference'' denotes the reference stimulus or auxiliary stimulus.  Star (*) marks that the model is established by ourselves according to original papers. ``Oracle'' means oracle phonetic feature.}
	\begin{tabular}{|c|c|c|c|c|}
		\hline
		 Task & Model&Reference & SI-SNRi & SDRi\cr
		\hline
         \multirow{2}{*}{BSS} & Conv-TasNet (PIT)* \cite{luo2019conv} &- &11.4487 & 11.7377 \\
                              & Multi-stage-AC* \cite{takahashi2020improving} &C& 14.4009 & 14.6527  \\
          
		\hline
          \multirow{4}{*}{SE}& $A_S$-ConvTasNet &$A_S$ & 11.2973&11.7457 \\
                                & AV-ConvTasNet* \cite{wu2019time} &V& 14.5356 & 14.7627 \\
                                & AC-ConvTasNet &C (Oracle)& 15.6915 & 15.9126 \\ 
                                & AVC-ConvTasNet &V+C (Oracle)& 14.8687 & 15.0889 \\\hline
          SE& VCSE & V+C& \textbf{15.8527} & \textbf{16.0800} \\\hline

	\end{tabular} \label{tab:result}
	\vspace{-10pt}
\end{table}

\section{Experimental results}
\label{sec:result}

We evaluate the system's performance using scale-invariant signal-to-noise ratio improvement (SI-SNRi) and signal-to-distortion ratio improvement (SDRi) \cite{vincent2006performance}, the higher the better for both metrics.

\subsection{Comparison with speech separation baselines}
We compare the VCSE with two speech separation algorithms, Conv-TasNet (PIT)~\cite{luo2019conv} and Multi-stage-AC model~\cite{takahashi2020improving} as shown in Table \ref{tab:result}, in which the latter model makes use of the self-enrolled contextual cues. The speech separation algorithms require knowing the number of speakers in the mixture.

The Conv-TasNet (PIT) model is a time-domain speech separation model that originally proposes the use of temporal convolutional neural network in estimating the intrinsic masks for every speaker. The Multi-stage-AC model is the two-stage method, which uses the Conv-TasNet (PIT) model to pre-separate the speech mixture and use an ASR model to extract the corresponding contextual features in the first stage. In the second stage, a TASNet incorporates the contextual feature and speech mixture to obtain the final estimated speech.

It is seen that the Multi-stage-AC model outperforms the Conv-TasNet (PIT) model due to the use of additional contextual cues. Our proposed VCSE outperforms the Multi-stage-AC model, as we utilize additional visual cues in pre-extracting the target speech in stage 1, which boosts the quality of the pre-extracted speech and self-enrolled contextual cues, therefore better higher signal quality for the final estimated speech. In addition, the VCSE network does not require prior knowledge of the number of speakers.

\subsection{Comparison with speaker extraction baselines}
We compare the VCSE network with other models using reference from different modalities as shown in Table \ref{tab:result}.
The $A_S$-ConvTasNet uses audio cues, i.e., a pre-recorded speech signal, to perform speaker extraction task. The audio cues mean an utterance taken from the target speaker that is not present in the current speech mixture. The AV-ConvTasNet follows ~\cite{wu2019time}, except that the visual encoder is not pre-trained on the lip reading task. Same as the VCSE, the AV-ConvTasNet model takes lip image sequence as reference to guide model extracting corresponding speech. The AC-ConvTasNet takes oracle phonetic embedding as reference for speaker extraction. This oracle phonetic embedding is comes from the E2EASR module which is applied on the clean speech. The AVC-ConvTasNet concatenates the representation of lip image sequence and oracle phonetic embedding as reference to extract target speech.

It's worth noting that $A_S$-ConvTasNet performs worse than Conv-TasNet (PIT) method. In our training set, there are about 5,000 speakers, and the total number of simulated utterances is 50,000. There isn't enough utterance from every speaker, which causes the network unable to distinguish each speaker's characteristic under the current amount of dataset. On the other hand, the visual/contextual cues are not affected by the number of speakers, showing the advantage of visual/contextual cues are more robust against dataset variations.

Our method significantly outperforms the $A_S$-ConvTasNet, as visual cues are much more robust against noise compared to the reference speech, not to mention that we also use the self-enrolled contextual cues.
The VCSE provides 1.3 dB improvement over AV-ConvTasNet that uses visual cues alone. This indicates the usefulness of our self-enrolled contextual cues. Our proposed VCSE even outperforms the AC-ConvTasNet which makes use of oracle contextual information, which shows the importance of the visual cues in speaker extraction. The result of AVC-ConvTasNet shows combining multiple modalities in a single model can not take full advantage of multimodal information. 
Our proposed VCSE takes both advantage of visual and contextual information through a combination of an audio-visual extraction module and an audio-contextual extraction module.

\subsection{Comparative study with variant of VCSE}
To verify the complementary effect of $s_t^{av}$, we design a variant of VCSE that does not utilize pre-extracted speech in the audio-contextual extraction module, referred to as VCSEv. Table \ref{tab:result2} shows the comparisons of VCSE and VCSEv. It shows that in the second stage, refining the $s_t^{av}$ is a better option compared to re-extraction from the speech mixture. The final target speech $s_t^{ac}$ of VCSE is derived from both visual and contextual cues, while $s_t^{ac}$ of VCSEv actually derives from self-enrolled contextual cues. This indicates our effective use of both the visual and contextual cues.

\renewcommand{\arraystretch}{2}
\begin{table}[tp]
	\centering
	\fontsize{7}{8}\selectfont
	\caption{Comparisons of VCSE and VCSEv in terms of SI-SNRi (dB) and SDRi (dB). ``V'' and ``C'' represent visual and contextual auxiliary information.  ``Input to AC" denotes the input variable to Audio-Contextual extraction module. }
	\begin{tabular}{|c|c|c|c|c|}
		\hline
		 Model  &Reference &Input to AC& SI-SNRi & SDRi\cr

		 \hline
		 
         VCSE  & V+C& $x_t,c_t,s_t^{av}$&   \textbf{15.8527}& \textbf{16.0800}\\\hline
         VCSEv  & V+C &$x_t,c_t$&15.3576&15.5834 \\ \hline

	\end{tabular} \label{tab:result2}
	\vspace{-10pt}
\end{table}

\section{Conclusions and future work}
\label{sec:conclusion}
In this paper, we propose a novel time-domain visual-contextual speaker extraction (VCSE) network to leverage visual cues and self-enrolled contextual cues for speaker extraction. Unlike the previous methods incorporating multiple modalities in a single model,  the VCSE network introduces visual information and contextual knowledge stage by stage to  take full advantage of different modalities. 
The experimental results show that our model achieves significant improvement over the previous methods. 
In the future, we are considering to investigate the complementary effect in depth between visual and contextual modalities.

\section{Acknowledgements}
\label{sec:Acknowledgements}
This research is supported by the National Key RD Program of China under Grant 2018YFB1305200, the internal project of Shenzhen Research Institute of Big Data under the Grant No. T00120220002, the Guangdong Provincial Key Laboratory of Big Data Computing under the Grant No. B10120210117- KP02, The Chinese University of Hong Kong, Shenzhen (CUHK-Shenzhen) and the University Development Fund, CUHK-Shenzhen, under the Grant No. UDF01002333 and UF02002333.


\vfill
\pagebreak

\bibliographystyle{IEEEtran}

\bibliography{refs}

\begin{thebibliography}{10}
\providecommand{\url}[1]{#1}
\csname url@samestyle\endcsname
\providecommand{\newblock}{\relax}
\providecommand{\bibinfo}[2]{#2}
\providecommand{\BIBentrySTDinterwordspacing}{\spaceskip=0pt\relax}
\providecommand{\BIBentryALTinterwordstretchfactor}{4}
\providecommand{\BIBentryALTinterwordspacing}{\spaceskip=\fontdimen2\font plus
\BIBentryALTinterwordstretchfactor\fontdimen3\font minus
  \fontdimen4\font\relax}
\providecommand{\BIBforeignlanguage}[2]{{%
\expandafter\ifx\csname l@#1\endcsname\relax
\typeout{** WARNING: IEEEtran.bst: No hyphenation pattern has been}%
\typeout{** loaded for the language `#1'. Using the pattern for}%
\typeout{** the default language instead.}%
\else
\language=\csname l@#1\endcsname
\fi
#2}}
\providecommand{\BIBdecl}{\relax}
\BIBdecl

\bibitem{bronkhorst2000cocktail}
A.~W. Bronkhorst, ``The cocktail party phenomenon: A review of research on
  speech intelligibility in multiple-talker conditions,'' \emph{Acta Acustica
  united with Acustica}, vol.~86, no.~1, pp. 117--128, 2000.

\bibitem{wang2008time}
D.~Wang, ``Time-frequency masking for speech separation and its potential for
  hearing aid design,'' \emph{Trends in amplification}, vol.~12, no.~4, pp.
  332--353, 2008.

\bibitem{tao2021someone}
R.~Tao, Z.~Pan, R.~K. Das, X.~Qian, M.~Z. Shou, and H.~Li, ``Is someone
  speaking? {E}xploring long-term temporal features for audio-visual active
  speaker detection,'' in \emph{Proc. of the 29th ACM Int. Conf. on
  Multimedia}, 2021, pp. 3927--3935.

\bibitem{ge2022spex}
M.~Ge, C.~Xu, L.~Wang, E.~S. Chng, J.~Dang, and H.~Li, ``L-spex: Localized
  target speaker extraction,'' in \emph{ICASSP 2022-2022 IEEE International
  Conference on Acoustics, Speech and Signal Processing (ICASSP)}.\hskip 1em
  plus 0.5em minus 0.4em\relax IEEE, 2022, pp. 7287--7291.

\bibitem{narayanan2014investigation}
A.~Narayanan and D.~Wang, ``Investigation of speech separation as a front-end
  for noise robust speech recognition,'' \emph{IEEE/ACM Trans. Audio, Speech,
  Lang. Process.}, vol.~22, no.~4, pp. 826--835, 2014.

\bibitem{yoshioka2018multi}
T.~Yoshioka, H.~Erdogan, Z.~Chen, and F.~Alleva, ``Multi-microphone neural
  speech separation for far-field multi-talker speech recognition,'' in
  \emph{Proc. IEEE Int. Conf. Acoust., Speech, Signal Process.}, 2018, pp.
  5739--5743.

\bibitem{wang2006computational}
D.~Wang and G.~J. Brown, \emph{Computational auditory scene analysis:
  Principles, algorithms, and applications}.\hskip 1em plus 0.5em minus
  0.4em\relax Wiley-IEEE press, 2006.

\bibitem{schmidt2006single}
M.~N. Schmidt and R.~K. Olsson, ``Single-channel speech separation using sparse
  non-negative matrix factorization.'' in \emph{Proc. INTERSPEECH},
  vol.~2.\hskip 1em plus 0.5em minus 0.4em\relax Citeseer, 2006, pp. 2--5.

\bibitem{yu2017permutation}
D.~Yu, M.~Kolb{\ae}k, Z.-H. Tan, and J.~Jensen, ``Permutation invariant
  training of deep models for speaker-independent multi-talker speech
  separation,'' in \emph{Proc. IEEE Int. Conf. Acoust., Speech, Signal
  Process.}, 2017, pp. 241--245.

\bibitem{hershey2016deep}
J.~R. Hershey, Z.~Chen, J.~Le~Roux, and S.~Watanabe, ``Deep clustering:
  Discriminative embeddings for segmentation and separation,'' in \emph{Proc.
  IEEE Int. Conf. Acoust., Speech, Signal Process.}, 2016, pp. 31--35.

\bibitem{zeghidour2021wavesplit}
N.~Zeghidour and D.~Grangier, ``Wavesplit: End-to-end speech separation by
  speaker clustering,'' \emph{IEEE/ACM Trans. Audio, Speech, Lang. Process.},
  vol.~29, pp. 2840--2849, 2021.

\bibitem{luo2019conv}
Y.~Luo and N.~Mesgarani, ``{Conv-TasNet}: Surpassing ideal time--frequency
  magnitude masking for speech separation,'' \emph{IEEE/ACM Trans. Audio,
  Speech, Lang. Process.}, vol.~27, no.~8, pp. 1256--1266, 2019.

\bibitem{ge2020spex+}
M.~Ge, C.~Xu, L.~Wang, E.~S. Chng, J.~Dang, and H.~Li, ``Spex+: A complete time
  domain speaker extraction network,'' \emph{arXiv preprint arXiv:2005.04686},
  2020.

\bibitem{han2020continuous}
C.~Han, Y.~Luo, C.~Li, T.~Zhou, K.~Kinoshita, S.~Watanabe, M.~Delcroix,
  H.~Erdogan, J.~R. Hershey, N.~Mesgarani \emph{et~al.}, ``Continuous speech
  separation using speaker inventory for long multi-talker recording,''
  \emph{arXiv preprint arXiv:2012.09727}, 2020.

\bibitem{9067003}
C.~Xu, W.~Rao, E.~S. Chng, and H.~Li, ``Spex: Multi-scale time domain speaker
  extraction network,'' \emph{IEEE/ACM Trans. Audio, Speech, Lang. Process.},
  vol.~28, pp. 1370--1384, 2020.

\bibitem{wang2018voicefilter}
Q.~Wang, H.~Muckenhirn, K.~Wilson, P.~Sridhar, Z.~Wu, J.~Hershey, R.~A.
  Saurous, R.~J. Weiss, Y.~Jia, and I.~L. Moreno, ``Voicefilter: Targeted voice
  separation by speaker-conditioned spectrogram masking,'' \emph{arXiv preprint
  arXiv:1810.04826}, 2018.

\bibitem{reentry}
Z.~Pan, R.~Tao, C.~Xu, and H.~Li, ``Selective listening by synchronizing speech
  with lips,'' \emph{IEEE/ACM Trans. Audio, Speech, Lang. Process.}, vol.~30,
  pp. 1650--1664, 2022.

\bibitem{aldeneh2021role}
Z.~Aldeneh, A.~P. Kumar, B.-J. Theobald, E.~Marchi, S.~Kajarekar, D.~Naik, and
  A.~H. Abdelaziz, ``On the role of visual cues in audiovisual speech
  enhancement,'' in \emph{Proc. IEEE Int. Conf. Acoust., Speech, Signal
  Process.}, 2021, pp. 8423--8427.

\bibitem{ephrat2018looking}
A.~Ephrat, I.~Mosseri, O.~Lang, T.~Dekel, K.~Wilson, A.~Hassidim, W.~T.
  Freeman, and M.~Rubinstein, ``Looking to listen at the cocktail party: A
  speaker-independent audio-visual model for speech separation,'' \emph{arXiv
  preprint arXiv:1804.03619}, 2018.

\bibitem{seg_pan}
Z.~Pan, X.~Qian, and H.~Li, ``Speaker extraction with co-speech gestures cue,''
  \emph{IEEE Signal Processing Letters}, 2022.

\bibitem{remley2017neuroscience}
D.~Remley, \emph{The neuroscience of multimodal persuasive messages: Persuading
  the brain}.\hskip 1em plus 0.5em minus 0.4em\relax Routledge, 2017.

\bibitem{ward2017enhanced}
N.~Ward, E.~Paul, P.~Watson, G.~Cooke, C.~Hillman, N.~J. Cohen, A.~F. Kramer,
  and A.~K. Barbey, ``Enhanced learning through multimodal training: evidence
  from a comprehensive cognitive, physical fitness, and neuroscience
  intervention,'' \emph{Scientific reports}, vol.~7, no.~1, pp. 1--8, 2017.

\bibitem{edelman1987neural}
G.~M. Edelman, \emph{Neural Darwinism: The theory of neuronal group
  selection.}\hskip 1em plus 0.5em minus 0.4em\relax Basic books, 1987.

\bibitem{rosenblum2008speech}
L.~D. Rosenblum, ``Speech perception as a multimodal phenomenon,''
  \emph{Current Directions in Psychological Science}, vol.~17, no.~6, pp.
  405--409, 2008.

\bibitem{pan2020multi}
Z.~Pan, Z.~Luo, J.~Yang, and H.~Li, ``Multi-modal attention for speech emotion
  recognition,'' in \emph{Proc. INTERSPEECH}, 2020, pp. 364--368.

\bibitem{chen2020multi}
Z.~Chen, S.~Wang, and Y.~Qian, ``Multi-modality matters: A performance leap on
  voxceleb.'' in \emph{Proc. INTERSPEECH}, 2020, pp. 2252--2256.

\bibitem{9755926}
Q.~Song, B.~Sun, and S.~Li, ``Multimodal sparse transformer network for
  audio-visual speech recognition,'' \emph{IEEE Transactions on Neural Networks
  and Learning Systems}, pp. 1--11, 2022.

\bibitem{wu2019time}
J.~Wu, Y.~Xu, S.-X. Zhang, L.-W. Chen, M.~Yu, L.~Xie, and D.~Yu, ``Time domain
  audio visual speech separation,'' in \emph{Proc. IEEE Autom. Speech Recognit.
  Understanding Workshop}, 2019, pp. 667--673.

\bibitem{pan2021muse}
Z.~Pan, R.~Tao, C.~Xu, and H.~Li, ``{MuSE}: Multi-modal target speaker
  extraction with visual cues,'' in \emph{Proc. IEEE Int. Conf. Acoust.,
  Speech, Signal Process.}, 2021, pp. 6678--6682.

\bibitem{pan2021usev}
Z.~Pan, M.~Ge, and H.~Li, ``{USEV}: Universal speaker extraction with visual
  cue,'' \emph{arXiv preprint arXiv:2109.14831}, 2021.

\bibitem{takahashi2020improving}
N.~Takahashi, M.~K. Singh, S.~Basak, P.~Sudarsanam, S.~Ganapathy, and
  Y.~Mitsufuji, ``Improving voice separation by incorporating end-to-end speech
  recognition,'' in \emph{Proc. IEEE Int. Conf. Acoust., Speech, Signal
  Process.}, 2020, pp. 41--45.

\bibitem{li2020listen}
C.~Li and Y.~Qian, ``Listen, watch and understand at the cocktail party:
  Audio-visual-contextual speech separation.'' in \emph{Proc. INTERSPEECH},
  2020, pp. 1426--1430.

\bibitem{petridis2018end}
S.~Petridis, T.~Stafylakis, P.~Ma, F.~Cai, G.~Tzimiropoulos, and M.~Pantic,
  ``End-to-end audiovisual speech recognition,'' in \emph{Proc. IEEE Int. Conf.
  Acoust., Speech, Signal Process.}, 2018, pp. 6548--6552.

\bibitem{afouras2018lrs3}
T.~Afouras, J.~S. Chung, and A.~Zisserman, ``Lrs3-ted: a large-scale dataset
  for visual speech recognition,'' \emph{arXiv preprint arXiv:1809.00496},
  2018.

\bibitem{vaswani2017attention}
A.~Vaswani, N.~Shazeer, N.~Parmar, J.~Uszkoreit, L.~Jones, A.~N. Gomez,
  {\L}.~Kaiser, and I.~Polosukhin, ``Attention is all you need,'' in
  \emph{Advances in neural information processing systems}, 2017, pp.
  5998--6008.

\bibitem{le2019sdr}
J.~Le~Roux, S.~Wisdom, H.~Erdogan, and J.~R. Hershey, ``{SDR}--half-baked or
  well done?'' in \emph{Proc. IEEE Int. Conf. Acoust., Speech, Signal
  Process.}, 2019, pp. 626--630.

\bibitem{vincent2006performance}
E.~Vincent, R.~Gribonval, and C.~F{\'e}votte, ``Performance measurement in
  blind audio source separation,'' \emph{IEEE/ACM Trans. Audio, Speech, Lang.
  Process.}, vol.~14, no.~4, pp. 1462--1469, 2006.

\end{thebibliography}

\end{document}